%% file: main.tex
\documentclass[sigconf]{acmart}
\AtBeginDocument{%
  \providecommand\BibTeX{{%
    \normalfont B\kern-0.5em{\scshape i\kern-0.25em b}\kern-0.8em\TeX}}}

\usepackage{colortbl} % 
\definecolor{Gray}{gray}{0.97}
\definecolor{Gray8}{gray}{0.9}
\definecolor{Gray18}{gray}{0.8}
\input{preameble}

\copyrightyear{2024}
\acmYear{2024}
\setcopyright{rightsretained}
\acmConference[BCIMM '24]{Proceedings of the 1st International Workshop on Brain-Computer Interfaces (BCI) for Multimedia Understanding}{October 28-November 1, 2024}{Melbourne, VIC, Australia}
\acmBooktitle{Proceedings of the 1st International Workshop on Brain-Computer Interfaces (BCI) for Multimedia Understanding (BCIMM '24), October 28-November 1, 2024, Melbourne, VIC, Australia}
\acmDOI{10.1145/3688862.3689108}
\acmISBN{979-8-4007-1189-3/24/10}

\begin{document}

\title{Enhancing End-to-End Autonomous Driving Systems Through Synchronized Human Behavior Data}

% \affiliation{%

\author{Yiqun Duan}
\email{yiqun.duan-1@uts.edu.au}
\orcid{0000-0003-1517-994X}
\affiliation{%
  \institution{HAI Centre, AAII, University of Technology Sydney}
  \city{Sydney}
  \state{NSW}
  \country{AU}
}

\author{Zhuoli Zhuang}
\email{zhuoli.zhuang@uts.edu.au}
\orcid{0009-0008-5088-3370}
\affiliation{
  \institution{University of Technology Sydney}
  \city{Sydney}
  \state{New South Wales}
  \country{Australia}
}

\author{Jinzhao Zhou}
\email{jinzhao.zhou@student.uts.edu.au}
\orcid{0000-0002-6620-604X}
%\authornotemark[2]
\affiliation{
  \institution{University of Technology Sydney}
  \city{Broadway}
  \state{New South Wales}
  \country{Australia}
}

\author{Yu-Cheng Chang}
\email{fred.chang@uts.edu.au}
\orcid{0000-0001-9244-0318}
\affiliation{
  \institution{University of Technology Sydney}
  \city{Broadway}
  \state{New South Wales}
  \country{Australia}
}

\author{Yu-Kai Wang}
\email{yukai.wang@uts.edu.au}
\orcid{0000-0001-8390-2664}
\affiliation{
  \institution{University of Technology Sydney}
  \city{Broadway}
  \state{New South Wales}
  \country{Australia}
}

\author{Chin-Teng Lin}
\email{chin-teng.lin@uts.edu.au}
\orcid{0000-0001-8371-8197}
\affiliation{
  \institution{University of Technology Sydney}
  \city{Broadway}
  \state{New South Wales}
  \country{Australia}
}

%% You do not have to enter your paper ID


\begin{abstract}
  This paper presents a pioneering exploration into the integration of fine-grained human supervision within the autonomous driving domain to enhance system performance.
    The current advances in End-to-End autonomous driving normally are data-driven and rely on given expert trials. However, this reliance limits the systems' generalizability and their ability to earn human trust. Addressing this gap, our research introduces a novel approach by synchronously collecting data from human and machine drivers under identical driving scenarios, focusing on eye-tracking and brainwave data to guide machine perception and decision-making processes. 
    This paper utilizes the Carla simulation to evaluate the impact brought by human behavior guidance. 
    Experimental results show that using human attention to guide machine attention could bring a significant improvement in driving performance. However, guidance by human intention still remains a challenge. 
    This paper pioneers a promising direction and potential for utilizing human behavior guidance to enhance autonomous systems. 
\end{abstract}

\begin{CCSXML}
<ccs2012>
 <concept>
  <concept_id>00000000.0000000.0000000</concept_id>
  <concept_desc>Do Not Use This Code, Generate the Correct Terms for Your Paper</concept_desc>
  <concept_significance>500</concept_significance>
 </concept>
 <concept>
  <concept_id>00000000.00000000.00000000</concept_id>
  <concept_desc>Do Not Use This Code, Generate the Correct Terms for Your Paper</concept_desc>
  <concept_significance>300</concept_significance>
 </concept>
 <concept>
  <concept_id>00000000.00000000.00000000</concept_id>
  <concept_desc>Do Not Use This Code, Generate the Correct Terms for Your Paper</concept_desc>
  <concept_significance>100</concept_significance>
 </concept>
 <concept>
  <concept_id>00000000.00000000.00000000</concept_id>
  <concept_desc>Do Not Use This Code, Generate the Correct Terms for Your Paper</concept_desc>
  <concept_significance>100</concept_significance>
 </concept>
</ccs2012>
\end{CCSXML}

%\ccsdesc[500]{Do Not Use This Code~Generate the Correct Terms for Your Paper}
%\ccsdesc[300]{Do Not Use This Code~Generate the Correct Terms for Your Paper}
%\ccsdesc{Do Not Use This Code~Generate the Correct Terms for Your Paper}
%\ccsdesc[100]{Do Not Use This Code~Generate the Correct Terms for Your Paper}

\keywords{Human-Guided Autonomous Driving; Brain-Computer Interfaces }



\maketitle

\input{intro}
\input{relatedworks}
\input{method}

\input{experiment}
%\input{supplementary}

% \begin{acks}
% \end{acks}

\bibliographystyle{ACM-Reference-Format}
\bibliography{main}

\end{document}

%% file: preameble.tex
\def \z {\mathbf{z}}

\def \I {\mathbf{I}}

\def \q {\mathbf{q}}

\newcommand{\bF}{\mathbf{F}} %
\newcommand{\bg}{\mathbf{g}}
\newcommand{\bh}{\mathbf{h}}

\newcommand{\bw}{\mathbf{w}}
\newcommand{\bx}{\mathbf{x}}

\newcommand{\nE}{\mathbb{E}}

\newcommand{\cD}{\mathcal{D}}

\newcommand{\cL}{\mathcal{L}}

\newcommand{\cW}{\mathcal{W}}
\newcommand{\cX}{\mathcal{X}}

\DeclareMathOperator*{\argmin}{argmin~}

\makeatletter
\DeclareRobustCommand\onedot{\futurelet\@let@token\@onedot}
\def\@onedot{\ifx\@let@token.\else.\null\fi\xspace}

\makeatother

%% file: intro.tex
\section{Introduction}
This position paper aims to pioneer the exploration of integrating granular human supervision into the burgeoning field of autonomous driving to enhance its performance. Presently, the majority of autonomous driving approaches, whether end-to-end or pipeline-based, rely heavily on expert trials. Such reliance is inadequate for considerations such as generalizability and earning human trust. This paper collects synchronous data from both human and machine driving scenarios to investigate this aspect.

The current trend in autonomous driving systems could be categorized into pipeline formations~\cite{song2020pip,wen2020scenario,huang2021bevdet,liu2022bevfusion} and End-to-End (E2E) approaches~\cite{tampuu2020survey,transfuser,zhang2022mmfn,shao2022safety}. 
These systems primarily transform raw sensory inputs into machine-readable representations using a variety of feature fusion techniques, such as BEV or range view. The goal is to develop autonomous driving models that can complete predetermined routes while safely navigating dynamic environments and complying with traffic rules, all through observational data-driven policies.

\iffalse
A common thread in these works is the adoption of Imitation Learning (IL) for policy development. This technique aims to replicate the behavior of an expert driver $\pi^{*}$, thereby creating a comparable autonomous policy $\mathbf{\pi}$. The training process involves creating a dataset $\cD = \{ (\cX^i, \cW^i) \}$ based on the expert's navigation of similar routes. Here, $\cX^i = \{(\bx_{im}^i, \bx_{Li}^i)_t \}_{t=1}^T$ represents combined image and LiDAR sensory data, while $\cW = \{ (x_t, y_t) \}_{t=1}^T$ captures the expert's trajectory of waypoints, with $x_t, y_t$ denoting the 2D coordinates in the ego vehicle's BEV space.
The primary objective is thus defined as $\argmin_{\pi} \mathbb{E}_{(\cX, \cW) \sim \cD} \left[ \cL_{wp} (\cW, \pi(\cX)) \right]$, where $\cL_{wp}$ indicates the waypoint loss, as detailed in Eq.~\ref{eq:waypointsloss_sup}. $\pi(\cX)$ here predicts waypoints based on the input $\cX$, guided by the learning policy $\pi$. This methodology ensures that autonomous systems like MaskFuser closely emulates expert driving patterns, effectively capturing the complexities and nuances of real-world driving scenarios.
\fi

These approaches mentioned above are predominantly data-driven, relying heavily on models learning from expert examples.
However, the pure data-driven formation also brings strong reliability to the data distribution. 
We suggest an augmentation of this methodology by incorporating fine-grained, immediate human labels as fine-tuning feedback to further enhance the robustness of the driving process. 
Our proposal involves a novel approach where humans and machines share the same driving scenarios. During these shared experiences, we aim to gather data from both human and machine drivers concurrently. 
This dual-data collection focuses on eye-gazing markers and brainwave data, offering a comprehensive view of the driving environment and decision-making processes.
%While existing learning methods for autonomous systems efficiently leverage simple, internet-based applications (e.g., extracting information from perception signals like classification and segmentation), and straightforward state-action decision-making (e.g., imitation learning), they still face considerable challenges. 
%These challenges often limit the performance of current imitation learning agents in exhibiting advanced intelligence.
Incorporating human-guided data into the autonomous system is intuitively right as humans still perform better ability while dealing with a lot of scenarios. 
The utilization of human guidance could leverage human superior intelligence in complex scenarios (eg. correct attention in complex scenarios) intuitively benefiting from expert guidance in the learning process. 

\begin{figure*}[hbpt]
    \centering
    \includegraphics[width=0.9\textwidth]{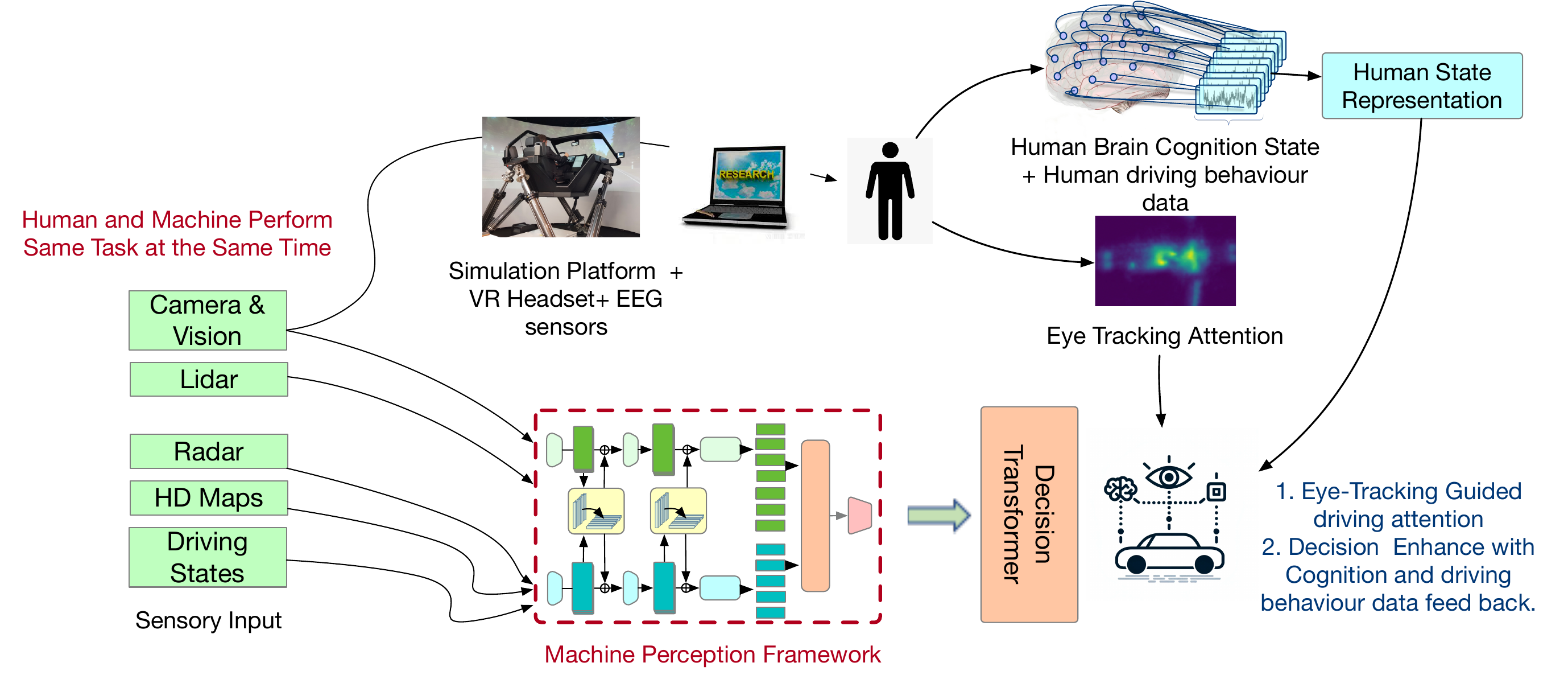}
    \caption{Visual schema of human enhanced autonomous driving. This model subjects both the machine and the human to identical driving scenarios along the same route. Throughout this process, we collect synchronous data on human behavior, including eye-tracking metrics, brainwave patterns, and brake signaling, capturing these elements in unison. }
    \label{fig:overview}
\end{figure*}

Introducing human guidance into the learning loop is a feasible and effective strategy, as evidenced by previous studies~\cite{schaal1999imitation,codevilla2019exploring,sun2024prompt,zhang2024whole}. However, this area has very limited research. 
Following the autonomous driving scenarios mentioned above, we propose the driving framework where the overall structure of our exploration is illustrated in Fig~\ref{fig:overview}.
It delineates the research into two main aspects of exploring how human guidance could enhance the autonomous driving system:  1) eye tracking attention during human driving and 2) human cognition data generated through the driving procedure.

We initially explore the concept of ``observing like a human" in Section~\ref{subsec:eye-tracking}. This involves analyzing eye movements during driving to uncover deeper insights into the cognitive processes and attentional focus of human drivers. Leveraging human attention as guidance, this approach seeks to address overlooked aspects by machines. It allows for the correction of instances where objects of potential significance are ignored by automated systems.
This human-guided attention is instrumental in highlighting objects that align with human empirical skills but may not be represented in expert training datasets. Such guidance is invaluable for machines, as it aids in recognizing and responding to critical elements in various driving scenarios, including sudden hazards or intricate traffic situations. By integrating these human-centric insights, we can significantly enhance the machine's perceptual and decision-making abilities in dynamic driving environments.

Secondly, we explore how the human cognition or behavior data could help guide the autonomous driving model in Section~\ref{subsec:cognition}. 
We explore the human cognition data to train an additional reward criticizer to provide additional feedback for additional imitation learning. 
To realize this goal, we first train a simple EEG wave classifier to recognize whether humans will decide to emergency break according to the current situation. 
Then, we feed the simultaneously collected EEG waves while human driving to this classifier and use the break signal as the additional feedback to the reinforcement fine-tuning of the driving model. 

Experimental results are conducted in Section~\ref{sec:exp_human-guide}, where we respectively explore how eye-tracking guidance and human cognition data guidance in Section~\ref{subsec:eye-tracking}. 
The results suggest leveraging human eye-tracking data as an auxiliary feedback mechanism in imitation learning has yielded positive improvements in driving performance. Specifically, this approach enhanced the driving score on the Carla Long-Set 6 dataset from 50.63 to 51.29. Conversely, the integration of cognition data did not result in a significant direct enhancement of the machine's driving score. 
At the position paper stage, we think this issue is reasonable because of two issues: 1) The recognition accuracy of the human brain data is still limited. This limited the guidance accuracy of human intention for driving. 2) The data quality is still relatively low, where the human driver's reaction speeds vary, and their decision-making capability varies when facing dangerous situations. 
This outcome suggests that the effective application of cognition data in autonomous driving remains an area ripe for further investigation. However, the effectiveness of human-observing data still outlines future research integrating human guidance in a collaborative human-machine driving context.
The contribution of this paper could be categorized into four folds. 

\begin{itemize}
    \item This position paper pioneers enhancing machine driving by two aspects 1) observing like human drivers, and 2) making decisions like human drivers. 
    \item This paper collects the parallel human cognition and behavior data of simultaneous machine and human driving. 
    \item Experimental results suggest effectiveness of introducing human-guidance into autonomous driving. 
\end{itemize}

%% file: relatedworks.tex
\section{Related Works}

\paragraph{End-to-End Autonomous Driving}

% background of leveraging human guidance in autonomous system 
Although rapidly developing, autonomous driving (AD) technology~\cite{shao2023safety}, it faces challenges in generalization and safety. To remedy this issue, a number of works have explored imitation learning (IL) and reinforcement learning methods to leverage human guidance for enhancing training efficiency and the safety of the learned of driving policy~\cite{codevilla2019exploring,shi2023efficient,tai2018socially,duan2024maskfuser,duan2024prompting}.  
% what are the existing method are proposed to incroprate human guidance
Current methodologies for integrating human guidance into autonomous driving (AD) systems primarily categorize based on their impact on the AD system into two distinct approaches: 1) leveraging human expert knowledge to train reliable driving policies, and 2) incorporating human oversight during model execution. Techniques such as Conditional Imitation Learning (CIL)~\cite{codevilla2018end}, and Generative Adversarial Imitation Learning (GAIL)~\cite{ho2016generative} rely on human expert trajectories for training the driving policy model. However, these methods face significant challenges, including the inefficiency of data collection and distribution shifts. Alternatively, a different set of approaches focuses on interpreting human biosignals during the evaluation phase, enabling the model to maneuver the car more safely under human guidance or feedback~\cite{gopinath2021maad}. Subsequent research has proposed strategies for employing real-time human guidance to enhance the safety and performance of the human-AI co-driving system~\cite{wu2023toward,wu2022prioritized,griffith2013policy,knox2011augmenting}. These strategies involve dynamic control transfer between human drivers and AI agents to facilitate timely intervention, thereby improving the co-driving experience. This evolution in approach underscores the importance of human-AI collaboration in enhancing the effectiveness and safety of autonomous driving systems. To combine the merits of the above methods, our proposed methods also make use of human biosignals to improve the training efficiency of AI models while also being used in driving interventions. 

% what human information are used in existing methods
\paragraph{Real-Time Human Guidance for Autonomous systems}
Current methods for introducing real-time human guidance in autonomous driving scenarios predominantly rely on head movement~\cite{rezaei2014look} and eye-tracking signals~\cite{zhou2021using,albadawi2022review}. However, these approaches, focus primarily on external indicators of attention and intent without considering the critical role of cognitive processes in driving. There are only limited works exploring the real-time monitoring of the cognitive process in the context of driving safety~\cite{sia2023eeg,wang2015eeg,cao2019multi,alyan2023blink}, however, they did not consider using the cognitive brain signal as guidance to the AD system. To address this gap, we propose an innovative research direction that integrates human cognitive signals alongside eye-tracking data to provide a more comprehensive understanding of driver intentions. By combining these sources of information, we aim to achieve a more seamless integration of the driver's objectives and safety considerations with the performance of the autonomous driving (AD) system. This holistic approach promises to enhance the symbiosis between human drivers and AD technologies, paving the way for advancements in driving safety and efficiency.

% \paragraph{End-to-End Autonomous Driving}

%% file: method.tex
\section{Methodology}
\label{sec:human-guided_driving}

\subsection{Overview}\label{subsec:problemset}

\textbf{Problem Setup:} 
We follow the previous widely accepted setting of E2E driving~(\cite{toromanoff2020end,transfuser,zhang2022mmfn}) that the goal is to complete a given route while safely reacting to other dynamic agents, traffic rules, and environmental conditions. 
Thus, the goal is to learn a policy $\mathbf{\pi}$ given observation.
We choose the \textbf{Imitation Learning (IL)} approach to learn the policy. The goal is to obtain policy  $\mathbf{\pi}$ by imitating the behavior of an expert $\pi^{*}$. 
Given an expert, the learning dataset $\cD = \{ (\cX^i, \cW^i) \}$ could be collected by letting the expert perform similar routes, where $\cX^i = \{(\bx_{im}^i, \bx_{Li}^i)_t \}_{t=1}^T$ denotes image and LiDAR sensory observations of the current state, and $\cW = \{ (x_t, y_t) \}_{t=1}^T$ denotes the expert trajectory of waypoints. 
Here, $x_t, y_t$ denotes the 2D coordinates in ego-vehicle (BEV) space.
Thus, the learning target could be defined as in Eq.~\ref{eq:target}.
\begin{equation}\label{eq:target}
 \argmin_{\pi} \nE_{(\cX, \cW) \sim \cD} \left[ \cL_{wp} (\cW, \pi(\cX)) \right]
\end{equation}
where $\cL_{wp}$ is the waypoint loss defined in Eq.~\ref{eq:waypointsloss_sup}, and $\pi(\cX)$ is the predicted waypoints given observation $\cX$ through policy $\pi$ to be learned. 

\textbf{Formation:}
In this paper, the policy $\pi(\cX)$ is realized by the combination of a hybrid fusion network (Sec.~\ref{subsec:hybridfusion}) and the decision transformer (Sec.~\ref{subsection:decisiontransformer}), where the fusion network transfer multi-modality sensory inputs $\cX$ into semantic tokens $\bF_s$, and decision transformer predicts future goal points $\cW$ given $\bF_s$. 
The human guidance is injected by  
Then a PID Controller is applied on the waypoints $\cW$ decision and decomposed into practical control, ie., steer, throttle, and brake. 

\iffalse
Following the proposed HybridNet with machine semantic representation introduced in Chapter~\ref{chap4}, Section~\ref{subsec:hybridfusion}, we further introduce reinforcement fine-tuning with human feedback to realize the human guidance injection. 
The overview of the framework is illustrated in Figure~\ref{fig:humanguidance}. 
To incorporate human guidance, we upgrade the single waypoint prediction GRU model mentioned in Section~\ref{subsec:waypoint} into the shown Decision transformer. It receives waypoints history trajectory, projected perception hidden state tokens, and both traffic query and cognition query and jointly predicts future waypoints $w_{t+1}, w_{t+2}, w_{t+3}, ...$, driving status (including velocity, heading, offset and bounding boxes), eye-tracking attention, traffic status (including traffic light, stop sign and junction), and human cognitive breaking intention. The direct logic is to inject the two aspects of human guidance by letting the model jointly predict the human attention and the cognition intention of the break action during imitation learning.  We will respectively give technical details as follows. 
\fi

\begin{figure*}[hbpt]
    \centering
    \includegraphics[width=0.78\textwidth]{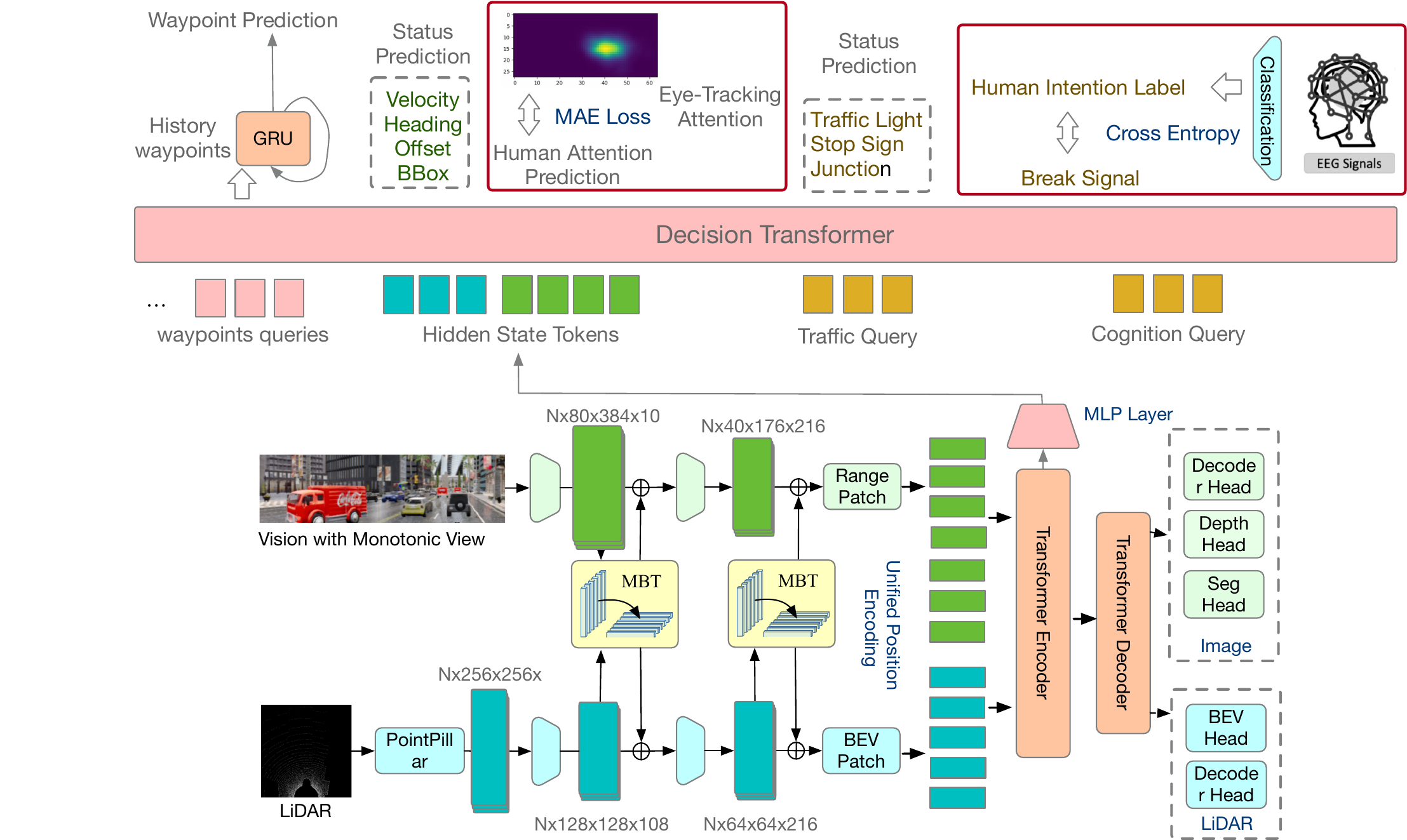}
    \vspace{-5pt}
    \caption{Framework of injecting human guidance into the autonomous system, taking autonomous driving as an example. The lower part is the hybrid fusion transformer encoder defined in Section~\ref{subsec:hybridfusion}.
    The learned machine state is fed into a decision transformer to come out with the final driving waypoint prediction. 
    The decision transformer is supervised by waypoints prediction GT and other safety constraints.
    The human guidance is injected by adding two human-behavior data supervision branches. The decision transformer is required to reconstruct human behavior (eye-tracking attention and brake intention) jointly with other targets. 
    %To incorporate human guidance, we upgrade the single waypoint prediction GRU model into the shown Decision transformer. It receives waypoints history trajectory, projected perception hidden state tokens, and both traffic query and cognition query and jointly predicts future waypoints $w_{t+1}, w_{t+2}, w_{t+3}$, driving status (including velocity, heading, offset and bounding boxes), eye-tracking attention, traffic status (including traffic light, stop sign and junction), and human cognitive breaking intention. 
    \label{fig:humanguidance} }
    \vspace{-5pt}
\end{figure*}

\subsection{Hybrid Fusion Transformer Encoder}
\label{subsec:hybridfusion}
For basic sensor fusion we utilize MaskFuser~\cite{duan2024maskfuser} as our main ourk. 
MaskFuser proposed a hybrid network shown in the lower part of Fig.~\ref{fig:humanguidance} that combines the advantages of \textit{early fusion} and \textit{late fusion}. 
The network consists of two stages. 
% Since behavior cloning is a straight forward modeling between observation and waypoints, the representation of feature is of vital importance. 
%Transfuser~\cite{transfuser} observed that pure geometrical fusion may under express the fusion results. 
%Yet, it ignores the spatial relations between semantic view and LiDAR space while apply transformer attention between CNN branches. 
%Thus, MaskFuser proposed a hybrid network shown in Fig.~\ref{fig:HybridNet} that combine the geometric fusion and semantic fusion. 
%The network is split into two stage. 

\textbf{Early Fusion:} At the first stage, we apply two separate CNN branches to extract shallow features respectively from monotonic image and LiDAR inputs. 
For the image branch, MaskFuser concatenates three front view camera inputs each with 60 Fov into a monotonic view and reshaped into shape $3\times160\times704$.
For the LiDAR branch, MaskFuser reprocesses the raw LiDAR input with PointPillar~\cite{pointpillars} into BEV feature with shape $33\times256\times256$.
Since the lower-level features still retain strong geometric relations, the separated encoder could extract tight local feature representation with fewer distractions. 
A novel monotonic-to-BEV translation (MBT) attention is applied to enrich each modality with cross-modality assistance. 
MBT attention translates both images and LiDAR features into BEV feature space and performs a more precise spatial feature alignment compared to previous element-wise approaches.

\textbf{Late Fusion:} At the second stage, the network respectively tokenizes~\cite{dosovitskiy2020image} feature maps from Image and LiDAR stream into semantic tokens, respectively denoted by green and blue in lower part of Fig.~\ref{fig:humanguidance}. 
% The tokens are added with a unified position embedding, and.
%where semantic fusion is applied. 
The late fusion is performed by directly applying a shared transformer encoder over the concatenated token representation. 
The shared encoder with position embedding could force the tokens from various modalities aligned into a unified semantic space. 
% Also, this ``modeling observation as language" formation enables the masked auto-encoder training defined in Sec.~\ref{subsec:cmmae}, which largely enhances the feature quality. 
Also, by treating multi-sensory observation as semantic tokens, we could further introduce masked auto-encoder training mentioned below. 

\subsubsection{Monotonic-to-BEV Translation (MBT)}\label{subsec:MBT}
MBT attention performs cross-modality attention more precisely by introducing human prior knowledge (BEV transformation). 
% As the MaskFuser uses monotonic view sensory inputs, 
Inspired by Monotonic-Translation~\cite{saha2022translating}, we model the translation as a sequence-to-sequence process with a camera intrinsic matrix. 
\begin{figure}[t]
  \centering
  \vspace{-10pt}
  \includegraphics[width=0.35\textwidth]{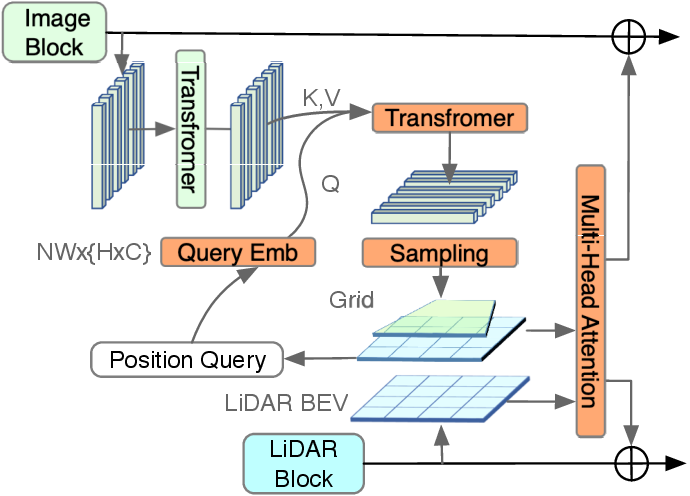}
  \caption{The structure of the MBT attention module, where the features from monotonic view are projected into BEV space through a sequence-to-sequence formation.  
  }
  \label{fig:mbt}
  \vspace{-10pt}
\end{figure}
The detailed structure of MBT attentions is shown in Fig.~\ref{fig:mbt}, where the feature map from image stream with shape $\bF_{im} \in \mathbb{R}^{N\times C\times H \times W}$ is reshaped along width dimension $W$ into image columns $\bF_c \in \mathbb{R}^{NW \times \{H \times C\}}$. 
The column vectors are projected into a set of mediate encoding $\{\bh_i \in \mathbb{R}^{ H \times C}\}_{i=1}^{NW}$ through a transformer layer with multi-head self-attention.  
%Based on the mediate encoding $\bh_i$, 
% A decoder transformer layer is applied to translate  
We treat mediate encoding of monotonic view $\bh_i$ which projects the key and value representing the range view information to be translated. 
\begin{equation}
    K(\bh_i) = \bh_i W_K, \quad V(\bh_i) = \bh_i W_V
\end{equation}
%The query embedding is 
% Since the MBT module is to translate the monotonic view into the BEV feature, we generate 
% A grid matrix is generated with the same shape as the BEV feature to be translated.  
A grid matrix is generated indicating the desired shape of the target BEV space.   
We generate position encoding $\{g_i \in \mathbb{R}^{r \times C }\}_{i=1}^{NW}$ along side each radius direction inside the grid with depth $r$. 
The grid position $g_i$ is tokenized into query embedding and query the $\bh_i$ with radius directions~\footnote{The radius coordinates are calculated by given camera intrinsic matrix, FOV, and prefixed depth length.} as defined in Eq.~\ref{eq:translate} and generate the translated feature map $\bF_{mbt}$.
\begin{equation}\label{eq:translate}
    \begin{aligned}
    \rm{s}(\bg_i, \bh_j) &= \frac{ Q(\bg_i) K(\bh_j)^T }{\sqrt{D}}, \quad Q(\bg_i) = \bg_i W_Q \\
    \bF_{mid} &= \sum _{i} \frac{\rm{exp}(\rm{s}(\bg_i, \bh_j))}{\sum _{j} \rm{exp}(\rm{s}(\bg_i, \bh_j))} V(\bh_i)
    \end{aligned}
\end{equation}
where $\rm{s}(\bg_i, \bh_j)$ is the scaled dot product~\cite{vaswani2017attention} regularized by dimension $D$. 
% The translated feature map $\bF_{mbt}$ is generated by query monotonic features with given polar ray position queries. 
Yet, since the monotonic view only suggests information inside certain FOV (Field of view), we apply a sampling process $\bF_{mbt} = \rm{P}(\bF_{mid})$ to sample points inside FOV decided by camera intrinsic matrix into BEV feature map $\bF_{mbt}$. 
The translated feature map $\bF_{mbt}$ and feature map from LiDar $\bF_{Li}$ is reshaped by flatten along width $W$ and depth $r$ into vectors and concatenate into sequence $\bF_{in} = \rm{cat}(\bF_{mbt}, \bF_{Li})$. 
The transformer layer is applied on $\bF_{in} \in \mathbb{R}^{N^\star \times C}$ to perform self multi-head attention~(\cite{vaswani2017attention}) between each token in $N^\star$ dimension.

\subsection{Decision Transformer}\label{subsection:decisiontransformer}
Inspired by the previous work InterFuser~\cite{shao2022safety}, we use a similar transformer decoder as the decision layer. 
The decoder follows the standard transformer architecture, transforming some query embeddings of size \( d \) using \( K \) layers of multi-headed self-attention mechanisms. 
Five types of input are designed: \( \{w_i\}^t_{i=1}\) previous waypoints query, \( R^2 \) density map queries (to query current vehicle status), human attention query, one traffic rule query, and one human intention query. 
These queries are concatenated into a sequence of tokens with shape $\q \in \mathbb{R}^{d \times N} $ which with the same shape with $\z$. 
To indicate the query order we add a standard positional embedding into the query $\q$ when doing the concatenation. 
In each decoder layer, we employ these queries to inquire about the desired information from the multi-modal multi-view features via the attention mechanism. 
More specifically, in each transformer layer, we treat perception state $\z$ as key \(\mathbf{K}\) and value \(\mathbf{V}\), and we treat the mentioned queries $\q$ as the query \(\mathbf{Q}\). 
This structure allows the decision transformer query perception state accordingly with different queries inside the sequence, resulting in the output independently decoded into waypoints, one density map, human attention prediction, traffic status, and the human intention by the following prediction headers.

\subsection{Basic Prediction Headers}
The transformer decoder is followed by five parallel prediction modules to predict the waypoints, the object density map, human attention, traffic rule, and human intention respectively.

\subsubsection{Waypoints Prediction}

For the waypoints prediction, following the mentioned waypoints prediction network defined~\cite{shao2023safety}, we take a single layer GRU to auto-regressively predict a sequence of three future waypoints \(\{w_{t+l}\}_{l=1}^3\). The GRU predicts the \( t +1 \)-th waypoints by taking in the hidden state from the \( t \)-th decoded waypoint embedding from the transformer decoder related to the waypoints queries, and previous inputs from the recorded $w_{t-2}, w_{t-1}, w_t$. Also to inform the waypoints GRU predictor of the ego vehicle’s goal location, we concatenate the GPS coordinates of the goal location at the beginning of the input sequence. 
More specifically, the loss function is defined in Equation~\ref{eq:waypointsloss_sup} as follows: 
\begin{equation}\label{eq:waypointsloss_sup}
    \cL_{wp} = \sum_{t=1}^T \left\|\bw_t - \bw_t^{gt}\right\|_1
\end{equation}
where the $\bw_t^{gt}$ is the ground truth from the expert route. 

\subsubsection{Density Map Status Prediction}
The density map prediction forces the model to learn to predict the current vehicle status on a density map. 
We basically follow the original setting of InterFuser~\cite{shao2022safety} as this is not our research target. 
%For the density map prediction, the corresponding \( R^2 \) length of \( d \)-dimensional embeddings from the transformer decoder are passed through a 3-layer MLP with a ReLU activation function to get a \( R^2 \times 7 \) feature map. We then reshape it into \( \mathbf{M} \in \mathbb{R}^{R \times R \times 7} \) to obtain the object density map. 
%The object density map is a \( R \times R \times 7 \) grid map with \( R \) rows, \( R \) columns, and 7 channels including 1 object probability channel and 6 object meta feature channels. 
%Given an object density map as a grid with dimensions \( R \times R \times 7 \), comprising 1 object probability channel and 6 object meta feature channels, we define the object density loss \( \cL_{\text{map}} \) to consist of a probability prediction loss \( \cL_{\text{prob}} \) and a meta feature prediction loss \( \cL_{\text{meta}} \):
%\begin{equation}
%    \cL_{\text{map}} = \cL_{\text{prob}} + \cL_{\text{meta}}
%\end{equation}
%The probability prediction loss aims at predicting the presence of objects within each grid cell. 
Please refer to appendix~A for detailed definitions of loss $ \cL_{map}$. 

\subsubsection{Traffic Rule Prediction}
For traffic rule prediction, the corresponding embedding from the transformer decoder is passed through a single linear layer to predict the state of the traffic light ahead, whether there is a stop sign ahead, and whether the ego vehicle is at an intersection. 
When predicting the traffic information \( \cL_{\text{tf}} \), we expect to recognize the traffic light status \( \cL_{l} \), stop sign \( \cL_{s} \), and whether the vehicle is at a junction of roads \( \cL_{j} \). All these three statuses are represented as a one-hot 0-1 label acquired from the CARLA simulator. These losses are simply calculated using cross entropy between prediction and ground truth as below:
\begin{equation}
    \cL_{\text{tf}} = \lambda_{l} \cL_{l} + \lambda_{s} \cL_{s} + \lambda_{j} \cL_{j} , 
\end{equation}
where \( \lambda \) balances the loss terms, which are calculated by binary cross-entropy loss.

\subsection{Human-Guidance Headers}
\subsubsection{Human Eye-Tracking Attention Prediction}
\label{subsec:eye-tracking}

We design the eye tracking attention queries with shape \( \mathbf{Q}_{\text{eye}} \in \mathbb{R}^{\frac{H}{16} \times \frac{W}{16} \times d} \), where the \( \frac{H}{16} \times \frac{W}{16} \) denotes the 16 times down-sampled range-view ratio of the camera, which has the exactly same ratio with the human attention ground truth map with ratio \( H \times W \). The query \( \mathbf{E}_{\text{eye}} \) is fed into the decision transformer by flattening \( \frac{H}{16} \times \frac{W}{16} \) into a fixed length and outputs the prediction embedding the same shape. Then we use a transpose convolution to upsample the embedding from shape \( \frac{H}{16} \times \frac{W}{16} \times d \) into human attention prediction \( \mathbf{E}_{\text{eye}} \) with shape \( H \times W \times 1 \). We define the eye tracking prediction loss \( \mathcal{L}_{\text{eye}} \) by calculating reconstruction loss between prediction \( \mathbf{E}_{\text{eye}} \) and ground truth human attention \( \bar{\mathbf{E}}_{\text{eye}} \):
\begin{equation}
    \mathcal{L}_{\text{eye}} = \lVert \mathbf{E}_{\text{eye}} - \bar{\mathbf{E}}_{\text{eye}} \rVert^2 ,
\end{equation}
where we use mean square error (MSE) loss to supervise the human attention predictor head.

\subsubsection{Human Intention Prediction}
\label{subsec:cognition}
Given the synchronized collected data human intention EEG data, we use a pre-trained classifier~\cite{duan2024dewave,zhou2024masked,zhou2024towards} to determine whether the human has the intention to break the car we respectively represent the EEG classified label as $\I_{EEG}$.
Meanwhile given the break behavior signal from the human driver's pedal as $\I_{brake}$, we propose to let the transformer jointly predict the human intention score. 
Similar to traffic rule prediction, we utilize a combined binary cross-entropy loss to supervise the decision transformer. The loss function is given below:

\begin{equation}
    \cL_{\text{hb}} = \lambda_\text{EEG} \cL_\text{EEG} + \lambda_\text{b} \cL_\text{brake} , 
\end{equation}
where the human intention is supervised by calculating the binary cross-entropy combination.

\subsection{Training Loss Combination}

The loss function is designed to simultaneously predict multiple targets including, waypoints (\( \cL_{pt} \)), object density map (\( \cL_{map} \)), human attention (\( \cL_{eye} \)), traffic rule (\( \cL_{tf} \)), and human break intention (\( \cL_{hb} \)).
\begin{equation}
    L = \lambda_\text{pt} \cL_\text{pt} + \lambda_\text{map} \cL_\text{map} + \lambda_\text{map} \cL_\text{map} + \lambda_\text{tf} \cL_\text{tf} + \lambda_\text{hb} \cL_\text{hb},
\end{equation}
where \( \lambda \) balances the three loss terms. Here, the waypoints (\( \cL_{pt} \)), object density map (\( \cL_{map} \)), and traffic rule (\( \cL_{tf} \)) consists of the pure data driving supervision for autonomous driving similar InterFuser~\cite{shao2022safety} and the two additional loss items human attention (\( \cL_{eye} \)) and human break intention (\( \cL_{hb} \)) indicates the human guidance. 

%% file: experiment.tex
\section{Experiments}
\label{sec:exp_human-guide}

\subsection{Data Collection}

\begin{figure}
    \centering
    \includegraphics[width=0.95\columnwidth]{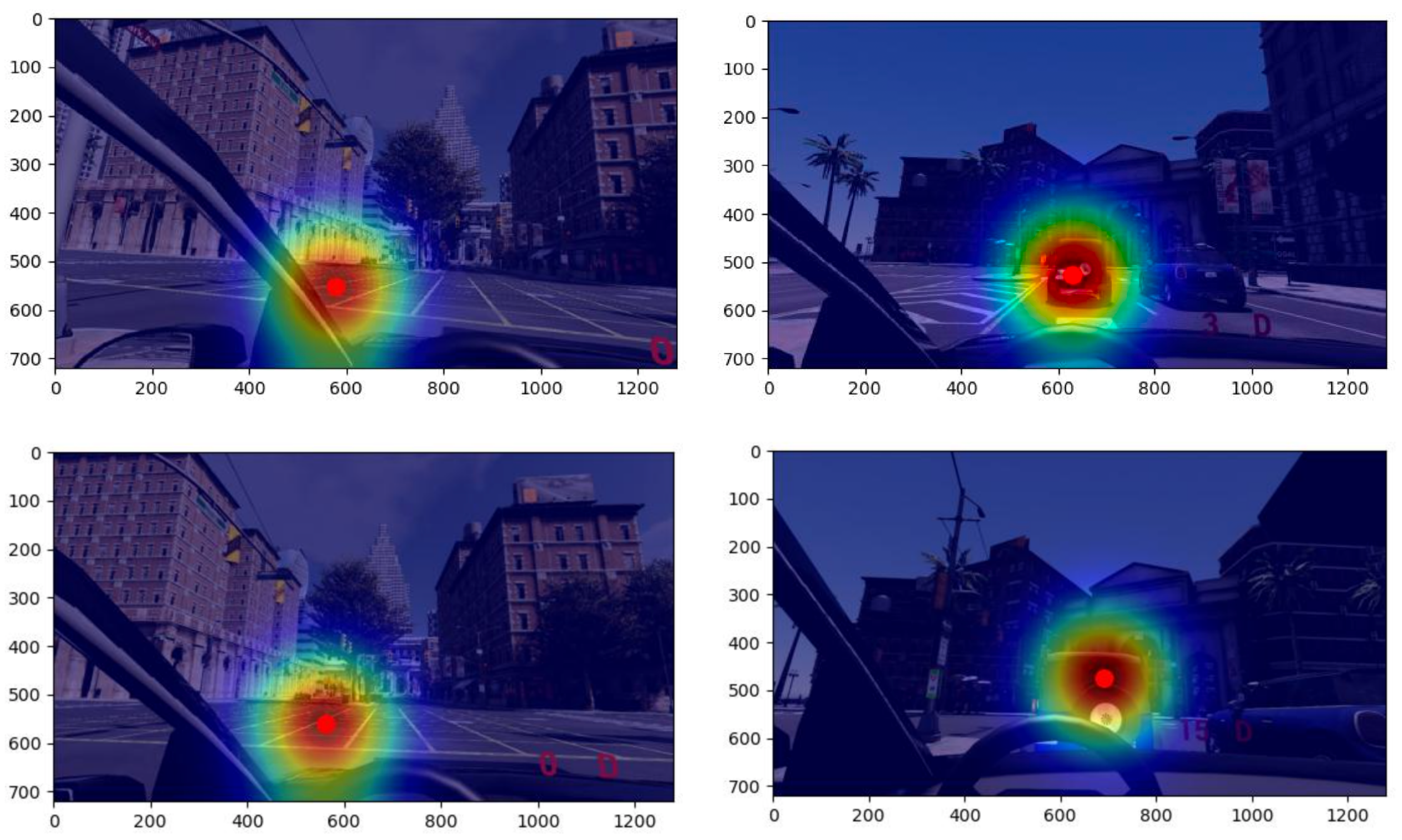}
    \caption{Visualization example of the collected combined human eye-tracking attention data.  \label{fig:human-eye}}
    \label{fig:enter-label}
\end{figure}

The data are collected by collecting data simultaneously from humans and machines driving under the same route and simultaneously collecting the human data including human eye-tracking, human brain waves, and braking behaviors. 
The simultaneous human collection process is realized by letting human subjects drive using the Logitech G920 driving force wheels, pedals, and driving chair. 
The human visual attention is collected by letting humans wear a HTC Vive Pro Eye virtual reality gear. 
We use the widely used CARLA simulator~\footnote{\href{https://github.com/carla-simulator/carla}{https://github.com/carla-simulator/carla}} version 0.9.13 as the driving simulator
The driving scenarios are projected into embodied perspective for the human driver. 
Thanks to the previous work DReyeVR~\footnote{\href{https://github.com/HARPLab/DReyeVR}{https://github.com/HARPLab/DReyeVR}}~\cite{silvera2022dreyevr}, we directly utilize the proposed modified CARLA simulator to complete the data collection. 
To visualize the human eye's attention, we provide examples of collected eye-tracking data in Figure~\ref{fig:human-eye}. 
Simultaneously, we collect the EEG data through a 64-channel collection device, where excluding the ground, reference, CB1, and CB2, the available count should be 60 out of 64.
We collect 12 human subjects using the default routes under Town 4 and Town 7 for human guidance. 

For the machine perception dataset, we employed a rule-based expert agent that adheres to the methodologies outlined in TransFuser~\cite{transfuser} and InterFuser~\cite{shao2022safety}. This agent was deployed across a diverse range of eight urban layouts and varying weather conditions, operating at a frequency of 2 frames per second. Through this process, we amassed an extensive expert dataset consisting of 3 million frames, which equates to approximately 410 hours of data. This dataset served as the foundation for the initial pretraining phase of our machine perception model.

\begin{figure}
    \centering
    \includegraphics[width=1.0\columnwidth]{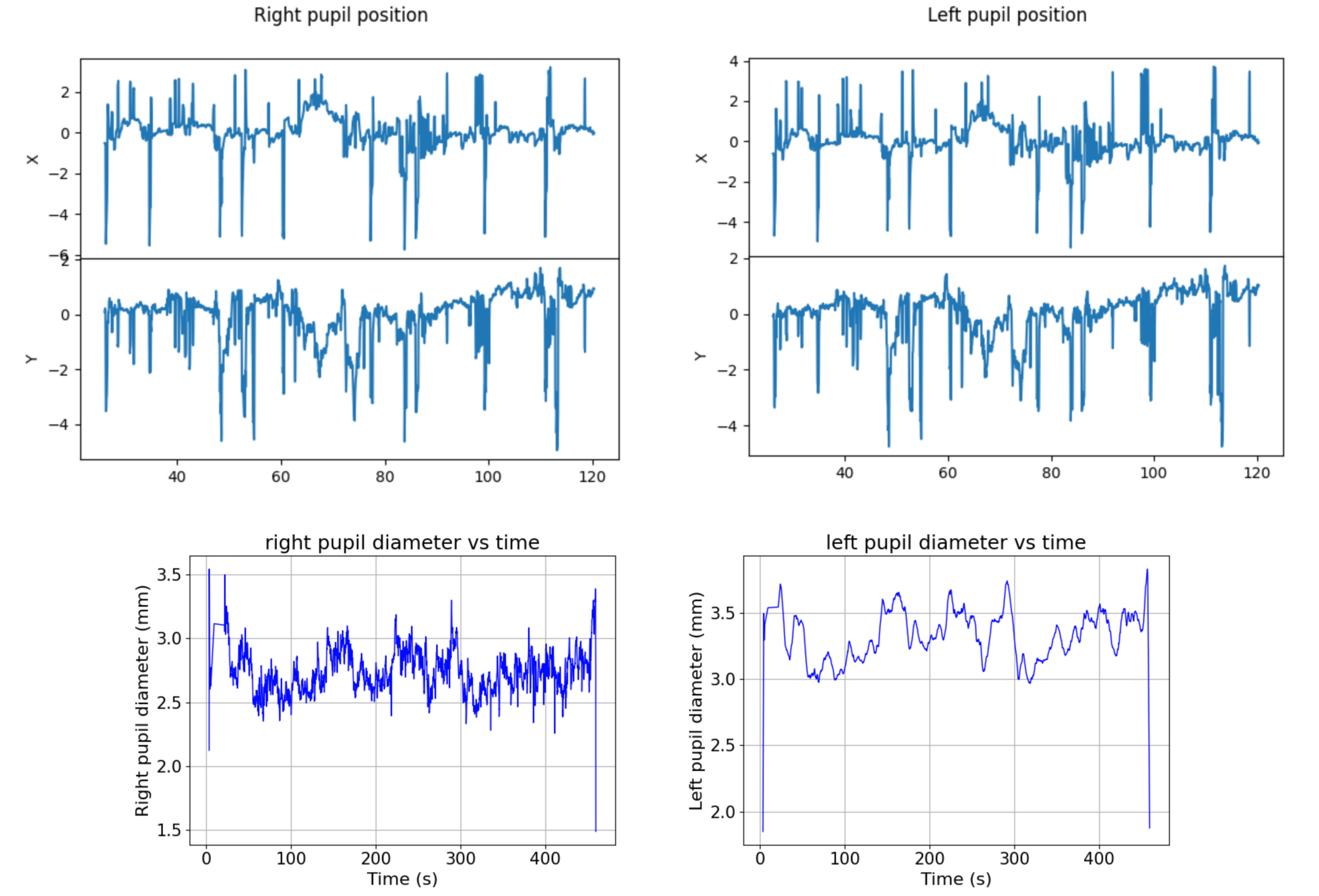}
    \caption{Example of the statistical distribution of left and right eye's pupil position and diameter size change during one episode.  \label{fig:eye-dist} }
\end{figure}

The eye-tracking is projected into 2D positions while training. 
For further illustration of the collected data distribution, we further visualize the statistical distribution of the eye data in Figure~\ref{fig:eye-dist}. 

\begin{table*}[hbpt]
\centering
\setlength\tabcolsep{5pt}
\caption{Driving Evaluation on LongSet6 with Human Guidance, where DT denotes decision transformer.\label{tb:longset6humanguidance}}
\centering
\resizebox{0.75\textwidth}{!}{
\begin{tabular}{cclcccc}
% \Xhline{0.9pt}
\toprule
\textbf{Human Guidance}& \textbf{DT} &      \textbf{Network}          & \textbf{DS $\uparrow$}          & \textbf{RC    $\uparrow$ $\scriptsize{(\%)}$}      & \textbf{IS$\uparrow$}       \\ \midrule
$\times$&$\times$ &TransFuser       & {46.95$\pm${5.81}} & {89.64$\pm${2.08}} & 0.52$\pm${0.08} \\
$\times$& $\times$ &MaskFuser        &  {49.05$\pm${6.02}}   &  {92.85$\pm${0.82}}  $\times$&   {0.56$\pm${0.07}}        \\ \midrule

$\times$& \checkmark & InterFuser       & {49.86$\pm${4.37}} & {91.05$\pm${1.92}} & {0.60$\pm${0.08}} \\
$\times$& \checkmark & MaskFuser \small{+ DT}  &  \underline{50.63$\pm${5.98}}   &  \textbf{92.89$\pm${0.80}}  &   {0.62$\pm${0.08}}      \\ \midrule

Eye-Attention & \checkmark & MaskFuser \small{+ DT}  &  \textbf{51.39$\pm${5.33}}   &  \underline{92.01$\pm${0.97}}  &   \textbf{0.65$\pm${0.09}}      \\ 
Intention & \checkmark & MaskFuser \small{+ DT}  &  {50.06$\pm${5.81}}   &  {90.97$\pm${1.80}}  &   \underline{0.63$\pm${0.09}}      \\ 
Eye-Attention + Intention & \checkmark & MaskFuser \small{+ DT}  &  {50.59$\pm${6.12}}   &  {91.39$\pm${0.80}}  &   \underline{0.63$\pm${0.09}}      \\ \midrule
%\rowcolor{Gray18} &  & Expert           &  75.83$\pm${2.45}   &   89.82$\pm${0.59}  &  0.85$\pm${0.03}      \\ \bottomrule
Eye-Attention + Fake Intention (GT) & \checkmark & MaskFuser \small{+ DT}  &  {51.28$\pm${6.17}}   &  {92.03$\pm${0.98}}  &   \underline{0.64$\pm${0.08}}      \\ \midrule
\rowcolor{Gray18} &  & Expert           &  75.83$\pm${2.45}   &   89.82$\pm${0.59}  &  0.85$\pm${0.03}      \\ \bottomrule
\end{tabular}
}
\end{table*}

\subsection{Implementation Details}
\label{exp:implementation}

The model training is divided into two stages, where stage 1 follows the normal autonomous model training, and stage 2 utilizes the human-guided data to further finetune the decision model. 

\textbf{Machine State Pretraining:}
For the feature extractor part, we follow the setting proposed in Section~\ref{exp:implementation}, the Hybrid Network, which employs cameras and LiDAR as dual modalities. Camera inputs are merged into a 120-degree FOV and reformatted to $(160, 704)$, while LiDAR data is transformed to a $256 \times 256$ BEV format~\cite{pointpillars}. For feature extraction, ImageNet-trained RegNet-32~\cite{xu2022regnet} is utilized on both image and BEV LiDAR data. Training includes angular augmentation by $\pm20$ degrees on LiDAR, akin to Transfuser~\cite{transfuser}, with corresponding label adjustments.
Inter-modal bridging transformers (MBTs) integrate features post-initial convolutions, at resolutions $(C_1, 40, 176)$ and $(C_2, 20, 88)$, where $C_1=72$ and $C_2=216$. The MBT comprises two transformer layers with 512 hidden dimensions and 4 heads. Following established polar ray grid sampling~\cite{saha2022translating}, MBT attends to varying depth ranges, translating to a real-world coverage of up to 30.5 meters.
Post-MBT, feature maps of $20\times88\times512$ and $32\times32\times512$ are sectioned into $4\times4$ patches, combined into a token sequence for a unified ViT encoder with four layers and 4 attention heads. This results in a semantic sequence of 174 tokens, maintaining the 512-dimensionality from early fusion.

For the decision transformer, the number of layers \( K \) in the transformer decoder and the transformer encoder is 6, and the feature dimension \( d \) is 256. The dimension of semantic token sequence output from the HybridNetwork is projected into 256 and fed into the decision transformer.  We train our models using the AdamW optimizer \cite{loshchilov2017decoupled} with a cosine learning rate scheduler \cite{loshchilov2016sgdr}. The initial learning rate is set to \( 5 \times 10^{-4} \times \frac{\text{BatchSize}}{512} \) for the transformer encoder \& decoder, and \( 2 \times 10^{-4} \times \frac{\text{BatchSize}}{512} \) for the encoders. The weight decay for all models is 0.07. All the models are trained for a maximum of 35 epochs with the first 5 epochs for warm-up \cite{loshchilov2016sgdr}. For data augmentation, we used random scaling from 0.9 to 1.1 and color jittering.

\textbf{Human-Guided Finetune} Upon completion of the pretraining with the 3 million frames derived solely from machine-generated data, we transitioned to a phase of targeted refinement. In this stage, the perception model's parameters were fixed, and we exclusively fine-tuned the decision-making module of the transformer. 
For the EEG wave classifier, we pre-train a ``human intention to brake" classifier based on the simultaneously collected EEG wave and the human behavior to brake label.
This refinement utilized a smaller, yet highly nuanced dataset comprising 12,000 frames of human-derived data. The objective was to integrate human decision-making nuances into the machine perception model, thereby enhancing its ability to interpret complex driving scenarios.
When we conduct the human-guided driving training, we use the initial learning rate as \( 1 \times 10^{-4} \times \frac{\text{BatchSize}}{512} \) for the decision transformer. The weight decay is kept the same as 0.07. 

\subsection{Evaluation with Human Guidance}

\noindent\textbf{Evaluation Benchmark:}
We directly keep the experimental settings the same as the pure machine driving evaluation on CARLA LongSet 6~\cite{transfuser}. 
We conduct our detailed ablation and comparison based on the Longeset6 Benchmark proposed by TransFuser~\cite{transfuser}, which chooses the 6 longest routes
per town from the officially released routes from the CARLA Challenge 2019 and shares quite a similarity with the official evaluation. 

\noindent\textbf{Quantative Evaluation}
For both online evaluation and offline evaluation, we follow the official evaluation metrics to calculate three main metrics, \textbf{Route Completion (RC)}, \textbf{Infraction Score (IS)}, and \textbf{Driving Score (DS)}.
The RC score is the percentage of route distance
completed. Given $R_i$ as the completion by the agent in route $i$, RC is calculated by averaging the completion rate $RC = \frac{1}{N} \sum_i^N R_i$. 
The IS is calculated by accumulating the infractions $P_i$ incurred by the agent during completing the routes. 
The driving score is calculated by accumulating route completion $R_i$ with infraction multiplier $P_i$ as $DS=\frac{1}{N} \sum_i^N R_i P_i$. 
We also calculate the detailed infraction statistical details according to the official codes. 
We report the evaluation with the human-guided fine-tuning models on these metrics in Table~\ref{tb:longset6humanguidance}.

It is observed from Table~\ref{tb:longset6humanguidance} that the integration of human-guided fine-tuning with the MaskFuser decision transformer (DT) leads to a noticeable improvement in Driving Score (DS). 
The improvement is mainly brought by human attention guidance, yet for the human intention guidance, we didn't observe a clear improvement. 
Here, the Eye-Attention guidance combined with the decision transformer achieves the highest mean driving score of $51.39$, indicating the effectiveness of incorporating human attention to guide the driving model. 
Furthermore, the addition of intention data slightly reduces the driving score under both settings. 
This phenomenon is still reasonable because of two reasons. 1) The human intention recognition is not as accurate at this stage as the supervise label. During our experiments, the human intention classifier has an accuracy between $60\%-70\%$ which is still compared low at this stage. 2) A long-existing problem appears that the simultaneously collected data has small time shifts between the human timestamp and the machine timestamp, this will lead to the wrong labels in rapidly changing situations. 
However, the improvement brought by the human attention guidance underscores the potential of leveraging nuanced human behavioral cues to enhance autonomous driving systems. 

\section{Limitation}
In the experimental section, although using eye-tracking data has a positive impact on autonomous driving performance, using human-intention data does not. We attribute this issue to two main factors: First, human intention relies on a pre-trained EEG recognition model, which can only achieve a $60-70\%$ accuracy rate in identifying people's danger or braking intentions, thereby introducing noise. Second, compared to the abundant machine-generated autonomous driving data, collecting driver intention data is relatively costly. We utilized driving data from only 12 individuals, which may not be sufficient to significantly influence the training of large models. Moreover, intention data, compared to human attention, provides sparser supervision, necessitating more supervised data for training. This aspect warrants further discussion in future work.

\section{Conclusion}
In this paper, we delve into the utilization of human behavioral data to improve autonomous driving performance. We explore harnessing insights from human drivers to enhance the driving system's capabilities by two aspects 1) observing like a human, and 2) decision like a human. To achieve this, we collected eye-tracking and brake $\&$ cognition data from 12 human subjects by letting machines and humans drive the same route. 
%By analyzing the human eye-tracking data, our approach guides the machine learning process to acknowledge objects or directions potentially ignored by the machine yet captured by human attention, effectively teaching the machine to 'observe' similarly to a human. Additionally, integrating a 'decision-making module' that mimics human cognitive processes, we utilize signals of human braking intent to warn the machine of imminent dangers. The findings from our experiments indicate that incorporating human eye-tracking data substantially elevates the efficacy of driving performance. 
%This method proposes to use human eye-tracking data to teach machines to recognize objects and directions that may be overlooked by algorithms but are noticed by humans, effectively enabling machines to 'observe' in a human-like manner. By incorporating a human cognition data module that simulates human cognition, we use human braking signals to alert machines to potential hazards. 
The experimental results indicate that guiding machine attention with human attention can lead to a clear improvement in performance. However, the experiments did not demonstrate that human cognition data could significantly enhance outcomes. Integrating granular human supervision into machine driving merits further in-depth investigation. Such an approach is beneficial for increasing the machine's trustworthiness to humans while making its decision-making processes more anthropomorphic.